\documentclass{article} 
\usepackage{url}
\pdfoutput=1

\usepackage{amsmath,amssymb,amsthm,amsfonts}
\usepackage{graphicx}
\usepackage{subcaption}
\def\lra#1{\left( #1 \right)}
\def\lrb#1{\left\{ #1 \right\}}
\def\lrc#1{\left[ #1 \right]}

\def\lraa#1{\bigl( #1 \bigr)}

\def\lrba#1{\bigl\{ #1 \bigr\}}

\def\bi#1{{\boldsymbol{#1}}}
\def\tpt{{\rm T}}

\def\myvec#1{\bi{#1}}

\def\myvect#1{\myvec{#1}^\tpt}

\def\figref#1{Figure \ref{#1}}

\title{A latent-observed dissimilarity measure}

\author{%
Yasushi Terazono\thanks{%
Department of Mathematical Informatics,
Graduate School of Information Science and Technology,
The University of Tokyo}
}

\date{}

\begin{document}
\maketitle

\begin{abstract}
Quantitatively assessing
relationships between latent variables
and observed variables
is important
for understanding and developing generative models and representation learning.
In this paper,
we propose latent-observed dissimilarity (LOD)
to evaluate the dissimilarity between
the probabilistic characteristics of latent and observed variables.
We also define four essential types of generative models
with different independence/conditional independence configurations.
Experiments using tractable real-world data
show that
LOD can effectively capture the differences between models
and reflect the capability for higher layer learning.
They also show that
the conditional independence of latent variables given observed variables
contributes to improving the transmission of information and characteristics
from lower layers to higher layers.
\end{abstract}

\section{Introduction}
\label{sec_intro}

Models with latent variables have been proposed and investigated
for explaining, understanding, or classifying observed data.
If a model is a generative model,
observed data are modeled to be as if they were generated
by latent variables
through parameterized probability distributions.
Popular criteria for learning generative models
include likelihood or posterior probability,
which both evaluate the probability of the given observed data
or parameters.
Another kind of criteria is mutual information.
Mutual information has been used to learn 
non-linear generative models~\cite{Pinchaud2011NIPS_mutual_information}
in which relationships between observed and latent variables
are directly evaluated.
It has also been used to learn linear
encoding (recognition) models%
~\cite{Baldi1995IEEE-NN_LNN,Obradovic1998NeuralComputation_infomax_ica}.






%

The relationships between observed and latent variables
have greater importance in more complex generative models,
e.g., deep learning models%
~\cite{Hinton2006science_autoencoder,Hinton2006NeuralComputation_DBN}.
In the pre-training of deep belief networks (DBNs),
one of the models or techniques of deep learning,
posterior samples of latent variables in the lower layer
are used as samples of observed variables in the next, higher layer.
For  successive layer learning to be possible,
latent variables should possess properties
that enable such learning.
It is crucial and fundamental for multiple layer learning theory 
to assess which observed variable properties 
are preserved, discarded, or modified in latent variables.
For this purpose, it is necessary to 
have good measures that capture
the capability of higher layer learning
and to know the configurations of models
suitable for higher layer learning.
%
%
Unfortunately, mutual information is not an adequate measure for this purpose.
The maximization of mutual information
is known to yield independent latent variables under certain conditions%
~\cite{Obradovic1998NeuralComputation_infomax_ica},
however, if latent variables are independent of each other,
successive learning exploiting their correlations becomes impossible.
%


%
In this paper, we propose
a novel measure
to capture the dissimilarity between latent and observed variables
in two-layer models.
We refer to the proposed measure as latent-observed dissimilarity (LOD).
The key idea is
to define a ``virtual-latent'' probability mass function (pmf) over observed variables,
using the conditionally expected information of latent variables.
This definition provides us with a new pmf
for which we can measure the dissimilarity from the original pmf.
The dissimilarity between these two pmfs
can be regarded as the dissimilarity between the latent and observed variables,
since the defined pmf reflects the conditionally expected information of latent samples,
while the original pmf reflects the self-information of observed samples.
We applied LOD to four essential types of two-layer models:
1) a single-latent-variable model (SL),
2) a multi-latent-variable model
whose latent variables are independent of each other (IL),
3) a multi-latent-variable model
whose latent variables are
conditionally independent given observed variables (CI),
and
4) a multi-latent-variable model
whose latent variables are
independent of each other
and conditionally independent given observed variables (ICI).
These four types cover
the major possible combinations of independence or conditional independence
in two-layer models.
In our experiments, LOD clearly reflected the difference between these four model types.
LOD was also shown to reflect the latent layer's capability for higher layer learning.
Our experiments also revealed that
the conditional independence of latent variables given observed variables,
particularly for CI models,
contributes to the improvement of higher layer learning,
improving LOD and the mutual information
between lower and higher layers.


\section{Latent-observed dissimilarity}
\label{sec_lod}




\subsection{Definition of LOD}
\label{ss_def_lod}

Let $p_{\rm G}\lra{X,Y}$ denote the probability mass function (pmf)
of a generative model
where $X$ denotes observed variables and $Y$ denotes latent variables.
When an observation $X$ is received,
its self information under a model $p_{\rm G}$ is given as $-\log p_{\rm G}\lra{X}$.
We first define the corresponding expected information
for latent variables.
Let $f\lra{X}$ denote
the expected information of $Y$ given $X$,
\begin{align}
f\lra{X}
& = E_{Y|X}\lrc{- \log {p_{\rm G}\lra{Y}}} \label{eq_fx_0} \\
& = - \textstyle\sum_Y p_{\rm G}\lra{Y|X}\log p_{\rm G}\lra{Y} \label{eq_fx}
\end{align}
where $f\lra{X}$ may be said to be the expected surprise of the latent layer given $X$,
while $-\log p\lra{X}$ is the surprise of the observed layer given $X$.

We then define a pmf $q\lra{X}$
based on $f\lra{X}$.
To measure the distance between some pmf and $f\lra{X}$,
preprocessing is necessary
because the function $f\lra{X}$ is not guaranteed to be a pmf.
Based on the fact that $f\lra{X}$ represents the expected information,
we define the following pmf,
\begin{gather}\label{eq_qx}
q\lra{X}
 = \frac{\exp\lra{- f\lra{X}}}{C},
\end{gather}
where
$C = \textstyle\sum_X \exp\lra{- f\lra{X}}$.
Let $\tilde{p}\lra{X}$ denote a data distribution.
That is, we assume
$\frac{1}{T}\sum_{t=1}^T g\lra{X\lra{t}}=\sum_X \tilde{p}\lra{X}g\lra{X}$
for any function $g$.
Using $q\lra{X}$, we define the dissimilarity between
the observed and latent variables for a dataset
using KL-divergence,
\begin{align}
{\rm LOD}\lra{X,Y}
&= D\lra{\tilde{p}\lra{X}||q\lra{X}}. \label{eq_SH_0}
\end{align}

\subsection{Characteristics of LOD}
\label{ss_char}


\paragraph{Single variable example.}
We now study the differences between LOD and mutual information
using single variable examples.

The proposed measure, LOD,
behaves differently from the mutual information of $X$ and $Y$.
When the joint probability of $X$ and $Y$ is defined by $p_{\rm G}\lra{X,Y}$,
the mutual information $I\lra{X;Y}$ between $X$ and $Y$ is
\begin{align}
I \lra{X;Y} 
& = \sum_X p_{\rm G}\lra{X} D_Y \lra{p_{\rm G}\lra{Y|X} || p_{\rm G}\lra{Y}}, \label{eq_MI_1}
\end{align}
where $D$ denotes the Kullback-Leibler divergence.
A more data-based evaluation is possible
if the data distribution $\tilde{p}\lra{X}$ is employed
\begin{gather}\label{eq_MI_2}
{\rm MI}\lra{X,Y}
 = \textstyle\sum_X \tilde{p}(X) D_Y \lra{p_{\rm G}\lra{Y|X} \| p_{\rm G}\lra{Y}}.
\end{gather}
We also refer to MI as (data-based) mutual information.

Consider the difference between LOD and MI in the simplest case.
Consider a model
consisting of a single observed variable and a single latent variable.
Let $X\in\lrb{x_1,x_2,\ldots,x_6}$ and $Y\in\lrb{y_1,y_2,\ldots,y_3}$.
Define the probabilities $\tilde{p} \lra{X=x_1}, \tilde{p} \lra{X=x_2}, \ldots, \tilde{p} \lra{X=x_6} $
as $1/21, 2/21, \ldots, 6/21$, respectively. 
For simplicity, we assume the mapping from $X$ to $Y$ to be deterministic,
so each $p_{\rm G}\lra{Y|X}$ is either $0$ or $1$.
%
%
From among all possible $p_{\rm G}\lra{Y|X}$ under this assumption,
let $p_1\lra{Y,X}$ denote the one that realizes the best LOD,
and let $p_2\lra{Y,X}$ denote the one that realizes the best MI.
The joint and marginal probabilities of $p_1$ and $p_2$
as well as the transformed probabilities $q\lra{X}$
are shown in Table \ref{tb_p1_p2_6}.
Note that since $p_1\lra{X}=p_2\lra{X}=\tilde{p}\lra{X}$ by assumption,
the log likelihood is maximized for both $p_1$ and $p_2$,
as $\sum_X \tilde{p}\lra{X}\log\tilde{p}\lra{X}
=\sum_X\tilde{p}\lra{X}\log p_1\lra{X}
=\sum_X\tilde{p}\lra{X}\log p_2\lra{X}$.
%
%
The scores of LOD and MI
are shown in Table \ref{tb_p1_p2_score_6}.

\begin{table}[tbp]
\caption{Joint and marginal probabilities of $p_1$ and $p_2$, and transformed probabilities $q\lra{X}$.
Top: the best similarity assignment. 
Bottom: the best mutual information assignment.
 Note that $a=1/21$, $b=1/42$.}
\label{tb_p1_p2_6}
\vspace{-2mm}
\begin{center}
\begin{tabular}{c|cccccc|c}
$p_1(X,Y)$ & $x_1$ & $x_2$ & $x_3$ & $x_4$ & $x_5$ & $x_6$ & $p_1 (Y)$ \\ \hline
$y_1$ & $a$ & $2a$ & $0$ & $0$ & $0$ & $0$  &  $3a$ \\ 
$y_2$ & $0$ & $0$ & $3a$ & $4a$ & $0$ & $0$ &  $7a$ \\
$y_3$ & $0$ & $0$ & $0$ & $0$ & $5a$ & $6a$ &  $11a$ \\ \hline
$p_1(X)$ & $a$ & $2a$ & $3a$ & $4a$ & $5a$ & $6a$ &  \\ \cline{1-7}
$q_1(X)$ & $3b$ & $3b$ & $7b$ & $7b$ & $11b$ & $11b$ &
\end{tabular}
\end{center}
%
\vspace{-1mm}
%
\begin{center}
\begin{tabular}{c|cccccc|c}
$p_2(X,Y)$ & $x_1$ & $x_2$ & $x_3$ & $x_4$ & $x_5$ & $x_6$ & $p_2 (Y)$ \\ \hline
$y_1$ & $a$ & $0$ & $0$ & $0$ & $0$ & $6a$  &  $7a$ \\ 
$y_2$ & $0$ & $2a$ & $0$ & $0$ & $5a$ & $0$ &  $7a$ \\
$y_3$ & $0$ & $0$ & $3a$ & $4a$ & $0$ & $0$ &  $7a$ \\ \hline
$p_2(X)$ & $a$ & $2a$ & $3a$ & $4a$ & $5a$ & $6a$ &  \\ \cline{1-7}
$q_2(X)$ & $7b$ & $7b$ & $7b$ & $7b$ & $7b$ & $7b$ &
\end{tabular}
\end{center}
\vspace{-2mm}
\end{table}

\begin{table}[tbp]
\caption{Scores for $p_1$ and $p_2$.
LOD: smaller is better.
MI: larger is better.
}
\vspace{-3mm}
\label{tb_p1_p2_score_6}
\begin{center}
\begin{tabular}{c|cc}
 & $p_1$ & $p_2$  \\ \hline
 LOD & $0.0137$ & $0.129$ \\
 MI & $0.983$ & $1.10$ \\
\end{tabular}
\end{center}
\vspace{-4mm}
\end{table}

From these results,
we can confirm the differences between
the minimum LOD model and the minimum MI model.
The model $p_1 \lra{X,Y}$ that minimizes ${\rm LOD}$
provides a $q_1\lra{X}$ that has a distribution similar to $\tilde{p}\lra{X}$.
The model $q_2\lra{X}$ from $p_2$ that minimizes MI is far from similar,
though the fact that MI is minimized in $p_2$ means
that knowing $Y$ in the $p_2$ model
reduces the uncertainty of $X$ more than in the $p_1$ model.



\paragraph{Sizes of latent/observed space.}
The proposed dissimilarity measure
LOD
achieves zero when $-\log\tilde{p}\lra{X}=f\lra{X}$.
However, 
there are other cases where LOD also achieves zero.
An illustrative case is the \textit{expanding} case
where the size of the latent space in the model
is an integer multiplication of the size of the observed space.
Let $K_A$ denote the total number of states
of observed variables, $K_A = \prod_i K_i$, and 
let $L_A$ denote the total number of states
of latent variables, $L_A = \prod_j L_j$.
Suppose $L_A=\alpha K_A$ by an integer $\alpha\ge 1$
and $p_{\rm G}\lra{X,Y}$ is defined as
\begin{gather}
p_{\rm G}\lra{Y=l|X=k}=
\begin{cases}
1/\alpha, & \text{if } \alpha\lra{k-1}+1\le l \le \alpha k .\\
0, & \text{otherwise}.
\end{cases}\nonumber
\end{gather}
This leads to $p_{\rm G}\lra{Y={\alpha\lra{k-1}+l}}=p\lra{X=k}/\alpha$
for $l=1,\ldots,\alpha$.
In this case,
$f\lra{X=k}=\log p_{\rm G}\lra{X=k} - \log \alpha$,
and hence
$q\lra{X=k}=p_{\rm G}\lra{X=k}$,
yielding ${\rm LOD}=0$.
The \textit{shrinking} case,
where $K_A=\beta L_A$ by an integer $\beta\ge 1$,
is also possible, which we shall omit the explanation.
%
%
The \textit{expanding}/\textit{shrinking}
cases show an invarance aspect of LOD,
which imply the potential advantage
of LOD as an optimization criterion
for the expansion and reduction of latent representation spaces.



\section{Models}
\label{sec_models}

In this section, the model types
used in our experiments (Section \ref{sec_exp}) are defined.
These model types differ in the independence or conditional independence
of their latent variables.
By comparing these models in our experiments,
we hope to determine which configurations
affect the relationships between observed, latent, and higher latent variables.

We consider the unsupervised learning of two-layer generative models
with four different configurations of latent and observed variables.
One of the layers is of observed, or manifest variables, $X$, and the other is of latent, or hidden variables, $Y$.
The stochastic variables $X$ and $Y$ are assumed to be finite and discrete, and $X$ and $Y$ may consist of multiple variables.
Let $N_x$ be the number of observation variables
and $N_y$ be that of latent variables.
In addition, let $K_i$, $i=1,\ldots,N_x$ be the number of states $X_i$ can take
and $L_j$, $j=1,\ldots,N_y$ be the number that $Y_j$ can take.
We denote a model probability by $p_{\rm G}\lra{X,Y}$.




The models and satisfied constraints
are summarized in Table \ref{tb_model_constraint}.
\begin{table}[tbp]
\caption{Models and satisfied constraints.}
\label{tb_model_constraint}
\vspace{-2mm}
\begin{center}
\begin{small}
\begin{sc}
\begin{tabular}{c|cccc}
\hline
 Constraint & SL & IL & CI & ICI \\
\hline
 $p\lra{X|Y}=\textstyle\prod_i p(X_i|Y)$ & \checkmark & \checkmark & \checkmark & \checkmark \\
 $p\lra{Y|X}=\textstyle\prod_j p(Y_j|X)$ & (\checkmark) &  - & \checkmark & \checkmark \\
 $p\lra{Y}=\textstyle\prod_j p(Y_j)$ & (\checkmark) &  \checkmark & - & \checkmark \\
\hline
\end{tabular}
\end{sc}
\end{small}
\end{center}
\vskip -0.1in
\vspace{-2mm}
\end{table}

\begin{figure}[htbp]
\begin{minipage}{0.49\hsize}
\centering
%
\includegraphics[width=35mm]{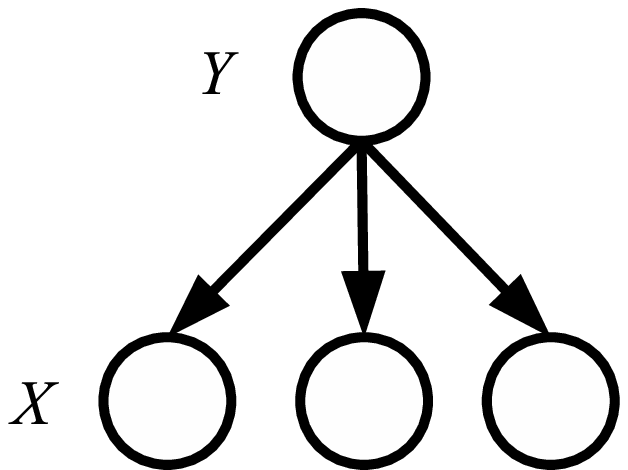}
\subcaption{a single-label model (SL).}
\label{fig_lcm}
\end{minipage}
\begin{minipage}{0.49\hsize}
\centering
%
\includegraphics[width=35mm]{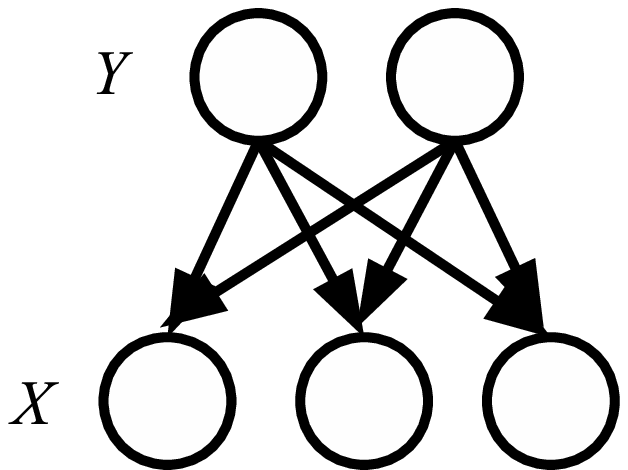}
\subcaption{an independent-label model (IL).}
\label{fig_mllcm}
\end{minipage}
\vspace{-2mm}
\caption{Bayesian network representations of single label and independent label models.}
\label{fig_lcms}
\vspace{-3mm}
\end{figure}

\paragraph{Single-label models (SL).}
The most simple of these models has a single latent variable
where each observed variable is conditioned only by the latent variable.
A Bayesian network representation of this model is shown in \figref{fig_lcm}.
This model is a type of mixture model
and is called a latent class or naive Bayes model
in different contexts.
The model assumes the conditional independence of $X$ given $Y$,
\begin{gather}\label{eq_x_ci}
p_{\rm G}\lra{X|Y} =  \textstyle\prod_{i=1}^{N_x} p(X_i|Y) .
\end{gather}
The joint probability of the model is
\begin{gather}\label{eq_lcm}
p\lra{X,Y} = \lrb{ \textstyle\prod_i p_{\rm G}\lra{X_i|Y} } p_{\rm G}\lra{Y}.
\end{gather}
We define each conditional probability by a conditional probability table,
\begin{gather}\label{eq_cpt_xoz}
p_{\rm G}(X_i=x|Y=y) = \Theta_{i,x}^y.
\end{gather}
We call this model the single label model (SL).
If $L$, the number of values $Y$ can take, is sufficiently large,
say $L \ge  \textstyle\prod_i K_i$,
then the model can realize any $p\lra{X}$.

\paragraph{Independent label models (IL).}
There are several ways to add more latent variables
to single-label models.
One is to add latent variables as indicated in \figref{fig_mllcm}.
Though the extension seems simple and straightforward
in the graphical representation,
the graph indicates the additional assumption that $Y$ is independent, that is, 
$p_{\rm G}\lra{Y}=\prod_j p_{\rm G}\lra{Y_j}$.
The joint probability is thus
\begin{gather}\label{eq_mllcm}
p_{\rm G}(X,Y) = \lrba{ \textstyle\prod_{i=1}^{N_x} p_{\rm G}(X_i|Y) } \lrba{ \textstyle\prod_{j=1}^{N_y} p_{\rm G}(Y_j)}.
\end{gather}
Models in this form have been proposed in different contexts, including
the probabilistic formulation of the quick medical reference network
(QMR-DT)~\cite{Shwe1990UAI_QMR,Jaakkola1999JAIR_variational_QMR},
and the partially observed bipartite network (POBN) used for the analysis of
transcriptional regulatory networks~\cite{Alvarez2011BMC_POBN}.
%
These models usually further restrict the form of probability.
In this paper, however,
we do not restrict $p_{\rm G}(X_i|Y)$ and $p_{\rm G}(Y_j)$ to some specific form.
We define each conditional probability by a conditional probability table,
$p_{\rm G}(X_i=x|Y_j=y) = \Theta_{i,x}^{j,y}$,
and $p(Y_j)$ is defined as
$p(Y_j=y)=\Phi_{j,y}$.
We call this model the independent label model (IL).



\begin{figure}[htbp]
\begin{minipage}{0.49\hsize}
\centering
%
\includegraphics[width=35mm]{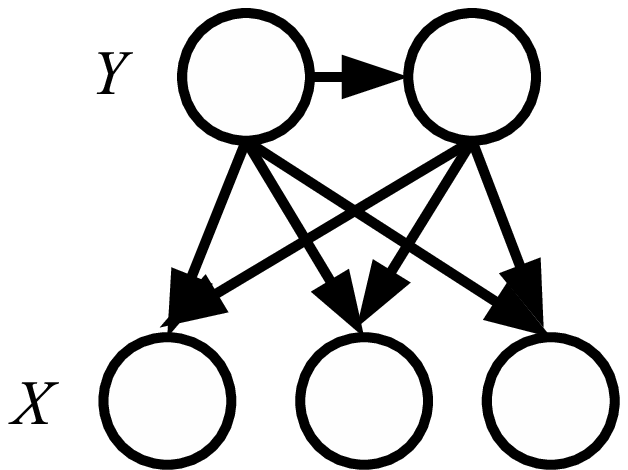}
\subcaption{Generative model, in which observed variables are conditionally independent given latent variables.}
\label{fig_ci_dec}
\end{minipage}
\begin{minipage}{0.49\hsize}
\centering
%
\includegraphics[width=35mm]{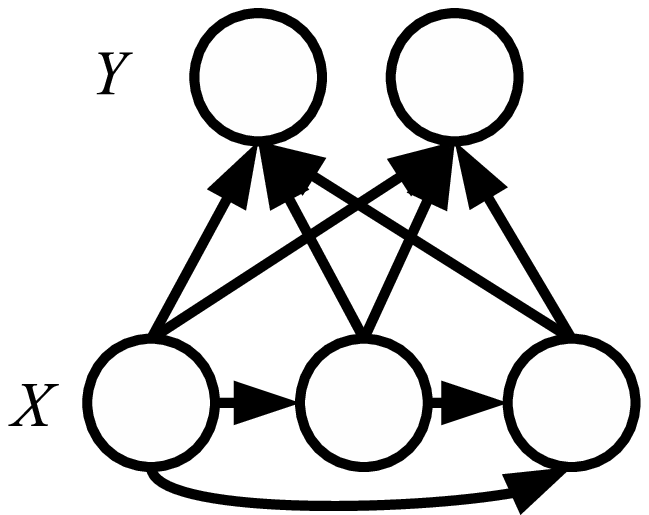}
\subcaption{Recognition model, in which latent variables are conditionally independent given observed variables.}
\label{fig_ci_enc}
\end{minipage}
\vspace{-2mm}
\caption{Two Bayesian network representations of a single probability model.}
\label{fig_ci}
\vspace{-2mm}
\end{figure}

\paragraph{Conditionally independent label models (CI).}
If independence is not assumed on multiple latent variables,
a model takes the form shown in \figref{fig_ci_dec}.
However, since $Z$ is latent and unsupervised learning is assumed,
models of this form are just equivalent to ``large'' single-label models.
A possible constraint
other than independence
is conditional independence of $Z$ given $X$.
\begin{align}
p_{\rm G}(X,Y) 
& = \lrba{ \textstyle\prod_i p_{\rm G}(X_i|Y) } p(Y) \label{eq_holo_01} \\
& = \lrba{ \textstyle\prod_j p_{\rm G}(Y_j|X) } p(X). \label{eq_holo_02}
\end{align} 
That is, latent variables are conditionally independent given observed variables,
while observed variables are conditionally independent given latent variables.
These two kinds of conditional independence are impossible to capture
in a single Bayesian network representation;
two Bayesian networks are necessary to illustrate two-way conditional independence.
\figref{fig_ci_dec} illustrates conditional independence in a generative model
and \figref{fig_ci_enc} illustrates it in a recognition model.
%
%
We call this model a conditionally independent label model (CI).

Joint probabilities satisfying these two-way constraints do exist.
An example class is that of the restricted Boltzmann machines (RBMs)%
~\cite{Smolensky1986Book_harmony_theory,Hinton2002NeuralComputation_PoE_CD}.
In an RBM, a joint probability of $X$ and $Y$ is defined as
$p_{\rm G}\lra{X,Y} = \frac{1}{D}
\exp
\lraa{
 -\myvect{a}X - \myvect{b}Y- Y^{\rm T}\myvec{W}X
}$,
where $D$ is the normalizing constant. This is often called a partition function.
Constraints \eqref{eq_holo_01} and \eqref{eq_holo_02} are consistently satisfied by RBMs.

If the generative part of a model is defined in the most general form,
that is, if it is parameterized as
$p_{\rm G}\lra{X_i=x_i|Y=y}=\Theta_{i,x_i}^{y}$ and $p_{\rm G}\lra{Y=y}=\Phi_y$,
the parameters $\Theta_{i,x_i}^{y}$ and $\Phi_y$
should be constrained to satisfy the recognition conditional independence \eqref{eq_holo_02}.
It is almost impossible to solve such constraints analytically;
however, a numerical, and perhaps approximate, satisfaction of the constraints
is possible through the framework of (stochastic) Helmholtz machines (HMs)
and the wake-sleep algorithm%
~\cite{Hinton1995Science_wake-sleep,Dayan1995NeuralComp_Helmholtz-Machine,Dayan1996NN_helmholtz-machine}.


%

\paragraph{Independent and conditionally independent label models}
If the independence and conditional independence constraints
are assumed simultaneously, the model satisfies
\begin{align}
p_{\rm G}(X,Y) 
& = \lrba{ \textstyle\prod_i p(X_i|Y) } \lrba{ \textstyle\prod_j p(Y_j) } \label{eq_ici_01} \\
& = \lrba{ \textstyle\prod_j p(Y_j|X) } p(X). \label{eq_ici_02}
\end{align} 
We call this model the independent
and conditionally independent label model (ICI).
Learning and (approximate) realization of this class of models
are also possible using the wake-sleep algorithm.

If $X$ and $Y$ are continuous and linearly mapped each other,
i.e., $X=\myvec{A}Y$ and $Y=\myvec{W}X$ where $\myvec{A}$ and $\myvec{W}$ are matrices,
the model represents independent component analysis (ICA)%
~\cite{Obradovic1998NeuralComputation_infomax_ica}.
In ICA, only $\myvec{W}$ is learned using some independence criterion.
The relationship between ICA and Helmholtz machines
has been investigated in, for example,
\cite{Xu1998Neurocomputing_Ying-Yang} and \cite{Ohata2003ICA_ICA_Helmholtz}.




\section{Experiments}
\label{sec_exp}

\subsection{Two-layer models}
\label{ss_exp_twolayer}

We first considered the two-layer models described in
Section \ref{sec_models}.
The models were trained on patches from images in the MNIST handwritten digits database%
\cite{LeCun_MNIST}.

\paragraph{Experimental settings.}
We preprocessed the images by quantizing them to three levels per pixel.
From each 28 $\times$ 28 pixel image,
a 2 $\times$ 2 pixel image patch was taken from a fixed location.
Thus, $N_x=4$, $K_1=\cdots=K_4=3$ for observed variable $X$.
We used all the training samples in the database,
so the number of samples $T$ was 60000.
Thus, a patch set consisted of 60000 samples of four observed variables,
where each variable is a ``trit'' (i.e., takes one of three values).
Eight non-overlapping locations were employed to yield eight such patch sets. 
To avoid the local minimum problem,
twenty trials were made for each patch set,
changing the initial parameters for the EM and the wake-sleep algorithm,
and the trial with the best log likelihood $(1/T)\sum_{t=1}^T \log p\lra{X\lra{t}}$
was chosen for each patch set.

The four kinds of models described in Section \ref{sec_models} were tested.
For the IL, CI, and ICI models,
$L_j$ $(j=1,\ldots,N_y)$ were fixed to two,
and $N_y$ was varied from one to six.
For the SL model,
$L$, the number of values $Y$ could take
were $2^1, 2^2, \ldots, 2^6$.
The SL and IL models were trained using the EM algorithm,
while the CI and ICI models were trained using the wake-sleep algorithm.

After learning,
we evaluated the learned models using
the following quantities:
a) log likelihood $(1/T)\sum_{t=1}^T \log p_{\rm G}\lra{X\lra{t}}$,
b) data-based mutual information MI \eqref{eq_MI_2},
c) the proposed dissimilarity measure LOD \eqref{eq_SH_0}.

To remove any large deviation
caused by different patch sets,
an offset removal procedure was performed as follows.
Let $V\lra{m,s,n}$ denote the raw evaluation values,
where $m$ denotes the model, $s$ denotes model size, and $n$ denotes patch set number.
1) The average of values of the smallest model in the series
was measured over the patch sets,
$V_a\lra{m} = \frac{1}{N} \sum_n V\lra{m,1,n}$.
2) From the evaluated values of a patch set model,
the value of the smallest model was subtracted,
$V_r\lra{m,s,n} = V\lra{m,s,n} - V\lra{m,1,n}$.
3) The average $V_a$ was then added back to $V_r$,
$V_q\lra{m,s,n} = V_a\lra{m} + V_r\lra{m,s,n}$.
Means and standard deviations were calculated for $V_q$
using $\frac{1}{N}\sum_n V_q\lra{m,s,n}$
and plotted.



\begin{figure*}[htb]
\begin{minipage}{0.49\hsize}
\centering
\includegraphics[width=0.9\columnwidth]{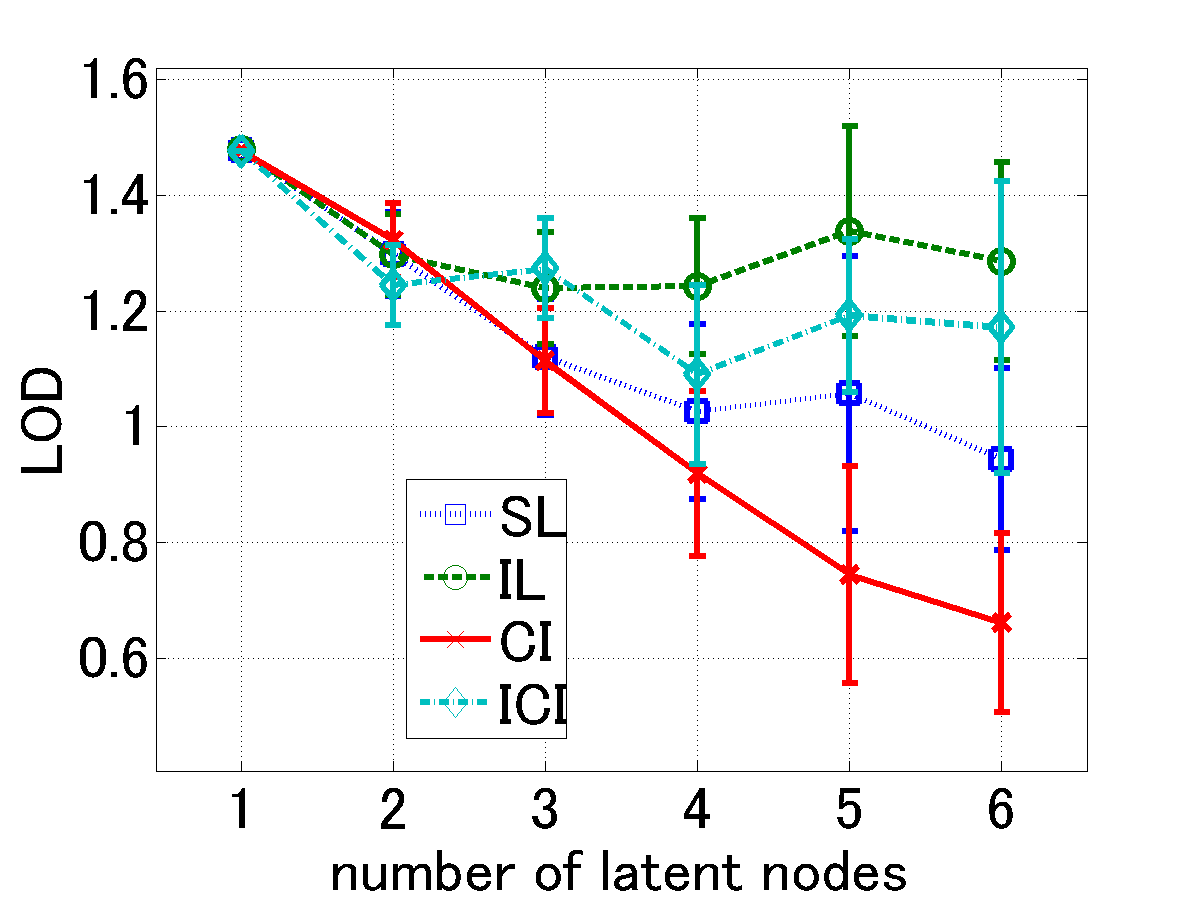}
\subcaption{LOD for different model configurations.}
\label{fig_aa_sh}
\end{minipage}
\begin{minipage}{0.49\hsize}
\centering
\includegraphics[width=0.9\columnwidth]{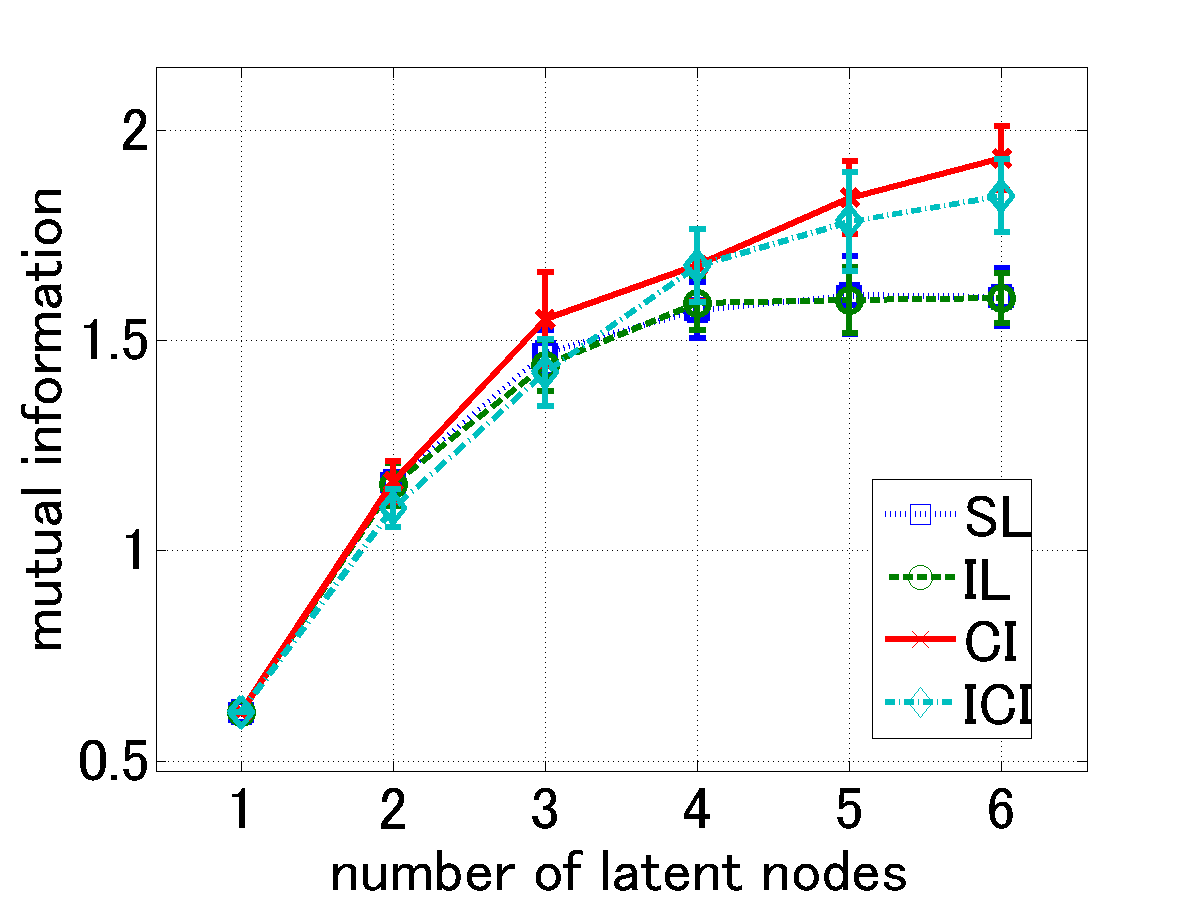}
\subcaption{Mutual information between latent and observed variables.}
\label{fig_aa_mi}
\end{minipage}
\begin{minipage}{0.49\hsize}
\centering
\includegraphics[width=0.9\columnwidth]{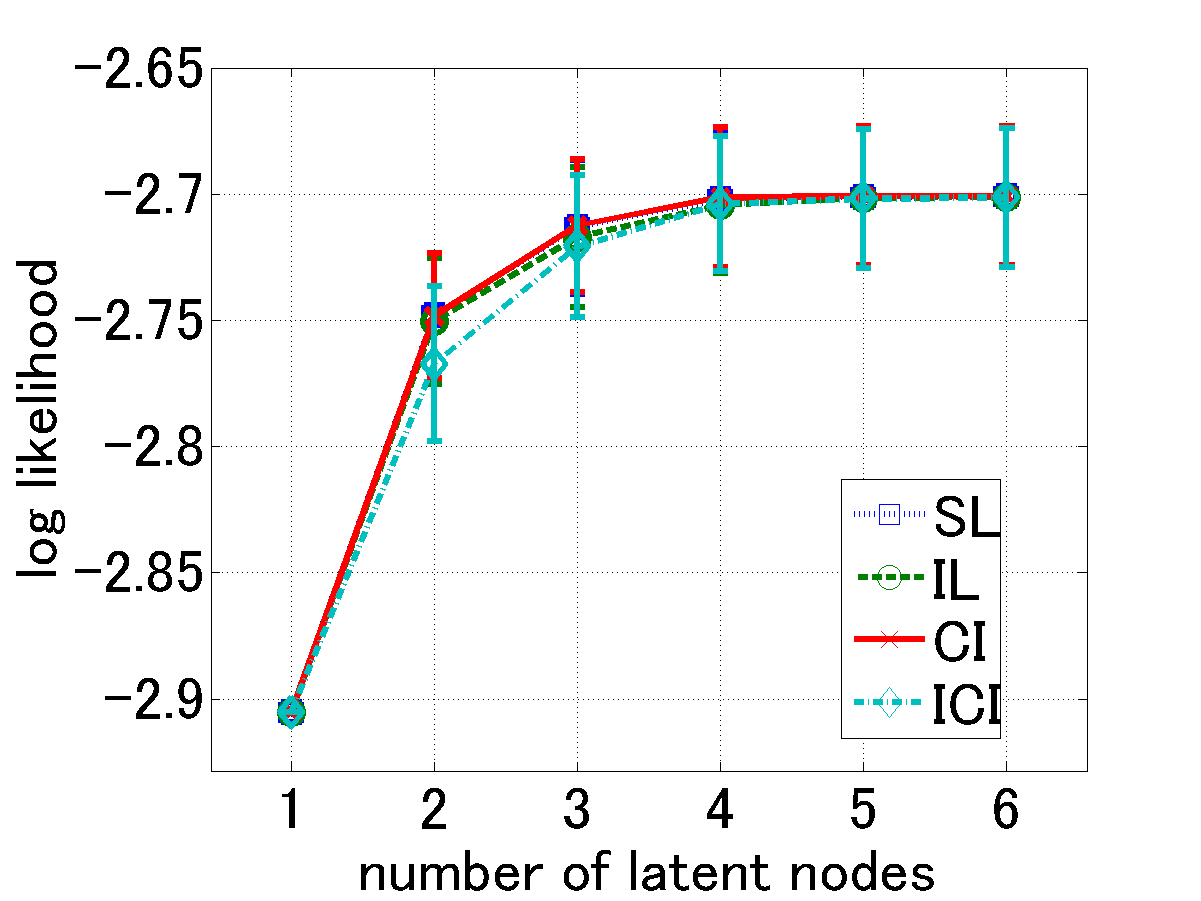}
\subcaption{Log likelihood for different model configurations.}
\label{fig_aa_ll}
\end{minipage}
\end{figure*}

\paragraph{Results: LOD.}
\figref{fig_aa_sh} shows LOD scores for the tested models.
CI has a lower LOD
than the other models for $N_y\ge 3$.
The graphs are, as a whole, decreasing for $N_y$,
but monotonic decrease holds only for CI.
For $N_y\ge4$, four types kept the order of
${\rm CI}<{\rm SL}<{\rm ICI}<{\rm IL}$.
This suggests that the conditional independence of latent variables
given observed variables improves LOD
because the essential difference between CI and SL as well as between ICI and IL
is the conditional independence.
%
Compared to MI and log likelihood,
LOD clearly captured the difference between model types.
The difference between LOD and log likelihood (\figref{fig_aa_ll})
indicates that the minimization of LOD
may lead to a model different from the maximum log likelihood model.
The incorporation of LOD into log likelihood as a regularization
may also be a future topic of discussion.


\paragraph{Results: Mutual Information.}
\figref{fig_aa_mi} shows the mutual information between the latent and observed variables
for the tested models.
All models show a monotonic increase of mutual information for $N_y$.
For $N_y\ge4$,
models appear to form two groups:
CI and ICI, and SL and IL.
The CI-ICI group took larger values than the SL-IL group, 
and in the CI-ICI group, CI was larger.
These phenomena can be explained as follows.
First, the conditional independence of the latent variables
contributed to a larger MI.
Secondly, the independence assumption on latent variables
did not affect MI as much as it affected LOD.
In fact, recalling the equivalence of ICA and mutual information maximization%
~\cite{Obradovic1998NeuralComputation_infomax_ica},
the independence assumption probably does not disturb the increase of mutual information.



\paragraph{Results: log likelihood.}
\figref{fig_aa_ll} shows the log likelihood of the tested models.
All models showed almost equally high likelihood for the same model size;
of these, ICI had a slightly lower value.
This is because ICI is the most restricted model among these four types
and the log likelihood was the objective of the optimization.

For LOD, MI, and log likelihood,
CI almost always yielded the best results.
This supports the incorporation of conditional independence
into models to improve the information transmission from the observed to latent variables
without penalizing the log likelihood too much.

\subsection{Learning of the higher (third) Layer}
\label{ss_higher}


Next, we performed learning of SL models
on top of the two-layer models learned in \ref{ss_exp_twolayer},
and evaluated
how the characteristics of the lower layers
are preserved or reflected in the higher layers.



\paragraph{Learning and evaluation procedures.}
Let us refer to the two-layer models learned in \ref{ss_exp_twolayer}
as the ``lower'' models, and denote their probability as $p_{\rm L}\lra{X,Y}$.
After learning these lower models,
a learning process similar to greedy layer-wise learning
in deep belief networks~\cite{Hinton2006NeuralComputation_DBN}
was carried out.
We applied each model's posterior distribution $p_{\rm L}\lra{Y|X}$
to the dataset used in \ref{ss_exp_twolayer}
to derive
$\tilde{p}\lra{Y}
:=\sum_X \tilde{p}\lra{X}p_{\rm L}\lra{Y|X}
=(1/T)\sum_t p_{\rm L}\lra{Y|X=\myvec{x}\lra{t}}$.
For the derived $\tilde{p}\lra{Y}$ of each model,
we learned a ``higher'' SL model,
$p_{\rm H}\lra{Y,Z}=\lraa{\prod_j p_{\rm H}\lra{Y_j|Z}}p_{\rm H}\lra{Z}$,
to maximize $\sum_Y \tilde{p}\lra{Y}\log p_{\rm H}\lra{Y}$,
where $Z$ denotes a set of the third layer latent variables.
The learning of $p_{\rm H}\lra{Y,Z}$ based on $\tilde{p}\lra{Y}$ is essentially equivalent to
the learning based on the samples $Y$ from $p_{\rm L}\lra{Y|X}$ for the dataset;
however, as model sizes are assumed to be small and tractability is ensured,
we can directly store and calculate $\tilde{p}\lra{Y}$
and do not need the actual samples from $p_{\rm L}\lra{Y|X}$.

The learning procedure yields
the higher two-layer SL models $p_{\rm H}\lra{Y,Z}$
on top of the lower two-layer model $p_{\rm L}\lra{X,Y}$.
We evaluated the correlations between
the lower model score $S\lra{X,Y}$ for $p_{\rm L}\lra{X,Y}$ and
the connected model score $S\lra{X,Z}$ for $p_{\rm C}\lra{X,Y,Z}$,
where the score was either LOD or MI.
The probability of a connected model $p_{\rm C}$ is defined by
\begin{gather}\label{eq_p_connected}
p_{\rm C}\lra{X,Y,Z}
= p_{\rm H}\lra{Z|Y}p_{\rm L}\lra{Y|X}\tilde{p}\lra{X}.
\end{gather}
In \eqref{eq_p_connected},
the lower and higher models are used as encoders,
because here we are focusing on how the higher layers preserve
 the characteristics of the lower layers
 and not on the generative properties of the models.

In the four lower model types,
higher model learning is impossible for the lower SL models
as they are, since SL models only have a single latent variable.
To make the learning of higher models possible,
the lower SL models were converted into multiple latent variable models as follows.
For the models whose number of states of $Y$ was $2^{m}$,
a corresponding model with $m$ binary latent variables
as in Figure \ref{fig_ci_dec} was defined.
Let $Y'=\lraa{Y'_1,\ldots,Y'_{m}}$ denote its latent variables.
The states of $Y$ can be mapped to the states of $Y'$
in a bijective (one-to-one and onto) manner.
Once such a bijection is determined,
the $m$-latent variables model and the SL model are
equivalent as generative models for $X$.
To determine a bijection for each lower model,
we first prepared twenty random bijections as the candidates.
For each bijection,
learning a higher SL model
with a single binary latent variable was performed,
and the bijection yielding the largest mean log likelihood
$\sum_Y \tilde{p}\lra{Y} \log p_{\rm H}\lra{Y}$
was selected from the twenty candidates.

\paragraph{Experimental settings.}
The experiment was configured as follows.
The number of datasets was eight, as in \ref{ss_exp_twolayer}, and
the lower models with $N_y=3,4,5,6$ were used.
For each lower model,
SL models with $K_z=2,3,\ldots,2^{N_y-2}$ were learned.
The number of the models used was thus
$8\times \lra{1+3+7+15} = 208$
for each lower model type (SL, IL, CI, and ICI).
Higher SL models were learned using the EM algorithm,
which we ran twenty times with different initial values, 
picking the run that gave the best log likelihood.
For the lower and higher models,
LOD and mutual information were evaluated using \eqref{eq_p_connected}
for between $X$-$Y$ and $X$-$Z$.

\begin{figure*}[htb]
\begin{minipage}{0.49\hsize}
\centering
\includegraphics[width=0.9\columnwidth]{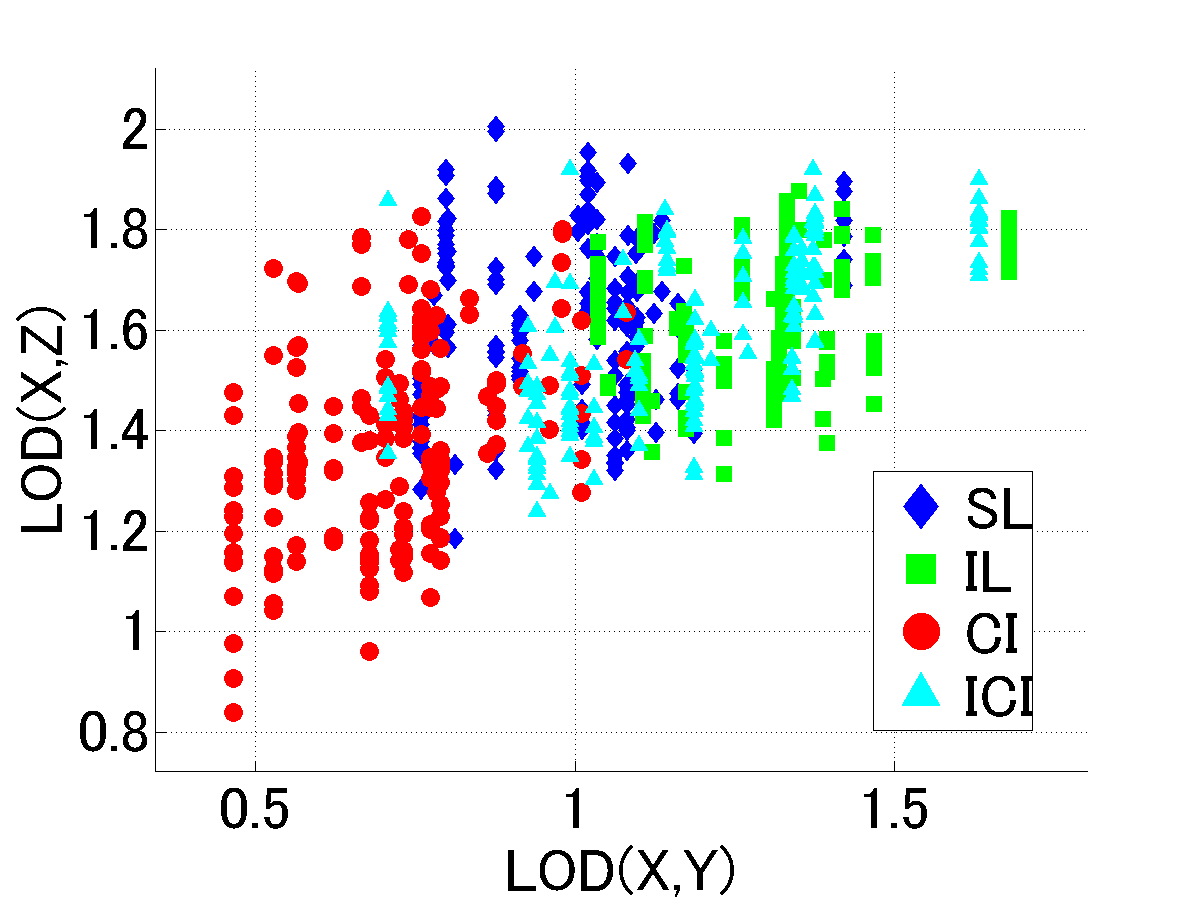}
\subcaption{${\rm LOD}\lra{X,Y}$ and ${\rm LOD}\lra{X,Z}$.}
\label{fig_lod1_lod3}
\end{minipage}
\begin{minipage}{0.49\hsize}
\centering
\includegraphics[width=0.9\columnwidth]{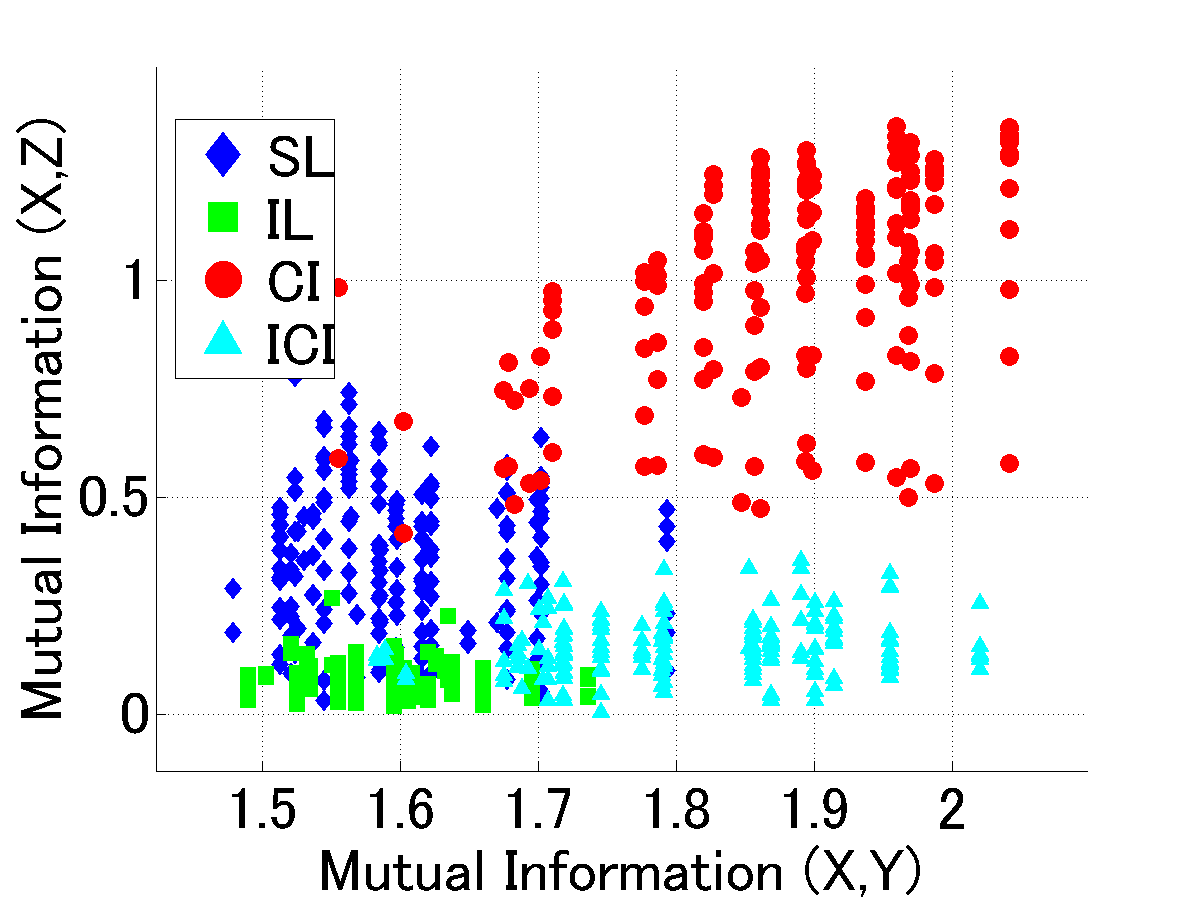}
\subcaption{${\rm MI}\lra{X,Y}$ and ${\rm MI}\lra{X,Z}$.}
\label{fig_mi1_mi3}
\end{minipage}
\caption{LOD and MI for different model configurations.}
\end{figure*}

\paragraph{Results.} Figure \ref{fig_lod1_lod3} shows
the relations between ${\rm LOD}\lra{X,Y}$ and ${\rm LOD}\lra{X,Z}$.
Figure \ref{fig_mi1_mi3} shows
the relations between ${\rm MI}\lra{X,Y}$ and ${\rm MI}\lra{X,Z}$.
Their correlations are shown in Table \ref{tb_corr_s1_s3}.

In Figures \ref{fig_lod1_lod3} and \ref{fig_mi1_mi3},
CI models achieved the lowest $X$-$Z$ dissimilarity
and the highest $X$-$Z$ mutual information among the four model types.
This indicates that latent variables encoded by CI models
keep more aspects of the information of the observed variables
than the other model types do.
From Table \ref{tb_corr_s1_s3},
the CI models had larger correlation coefficients than those of SL models for both LOD and MI. This relationship was also true for the ICI models and IL models.
The capability of $Y$ to provide information to $Z$
was improved by the incorporation of the conditional independence
of the latent variables given observed variables.

LOD for $\lra{X,Y}$ and $\lra{X,Z}$ showed
significant $\lra{p<0.05}$ correlations for all of the four model types,
whereas MI showed significant correlations only for CI and ICI models.
These results indicates that,
along with dissimilarity itself,
LOD also represents how well similarity can be transmitted to the higher layer,
whereas MI does not necessarily represent such a capability of transmission.
%


\begin{table}[htbp]
\caption{Correlations between ${\rm Score}\lra{X,Y}$ and ${\rm Score}\lra{X,Z}$.
Score is either LOD or MI.
$r$ means the correlation coefficient, and $p$ means the p-value.
See Figures \ref{fig_lod1_lod3} and \ref{fig_mi1_mi3} for the source data.}
\label{tb_corr_s1_s3}
\vspace{-5mm}
\centering
\begin{minipage}{0.95\hsize}
\vskip 0.15in
\centering
\begin{tabular}{c|cc|cc}
    & \multicolumn{2}{c|}{LOD} & \multicolumn{2}{c}{MI} \\
MODEL & $r$ & $p$ & $r$ & $p$ \\
\hline
(SL) & $0.157$ & $0.037$ & $-0.0664$ & $0.381$ \\
(IL) & $0.373$ & $<0.001$ & $-0.0586$ & $0.440$ \\
(CI) & $0.410$ & $<0.001$ & $0.510$ & $<0.001$ \\
(ICI) & $0.602$ & $<0.001$ & $0.156$ & $0.039$ \\
\end{tabular}
\end{minipage}
\vspace{-3mm}
\end{table}



\section{Conclusions}
\label{sec_conclu}

We proposed latent-observed dissimilarity (LOD),
a dissimilarity measure
between latent and observed variables
in generative models,
to evaluate the relationships between
latent and observed variables.
LOD compares the self-information of an observation
with the expected information of a latent layer
given that observation.
We numerically evaluated four types of two-layer models (SL, IL, CI, and ICI) using log likelihood,
mutual information, and LOD.
The results suggested an advantage of using LOD as a measure for multi-layer learning;
the LOD between observed and latent variables
had significant correlation with the LOD between observed and higher layer latent variables
for all four types of models,
while mutual information had significant correlation only for CI models.
The results also suggested the conditional independence of latent variables given observed variables
facilitates the transmission of a layer's characteristics to the higher layers.
This fact sheds new light on the advantages of conditional independence,
of which usually only its computational advantage is emphasized.

\section*{Acknowledgement}

This work was supported by MEXT KAKENHI Grant Number 23240019.




\begin{thebibliography}{10}

\bibitem{Alvarez2011BMC_POBN}
Angel Alvarez and Peter~J. Woolf.
\newblock Partially observed bipartite network analysis to identify predictive
  connections in transcriptional regulatory networks.
\newblock {\em BMC Systems Biology}, 2011.

\bibitem{Baldi1995IEEE-NN_LNN}
Pierre Baldi and Kurt Hornik.
\newblock Learning in linear neural networks: a survey.
\newblock {\em IEEE Transactions on neural networks}, 6(4):837--858, 7 1995.

\bibitem{Dayan1996NN_helmholtz-machine}
Peter Dayan and Geoffrey~E. Hinton.
\newblock Varieties of helmholtz machine.
\newblock {\em Neural Networks}, 9(8):1385--1403, 1996.

\bibitem{Dayan1995NeuralComp_Helmholtz-Machine}
Peter Dayan, Geoffrey~E. Hinton, Radford~M. Neal, and Richard~S. Zemel.
\newblock The helmholtz machine.
\newblock {\em Neural Computation}, 7:889--904, 1995.

\bibitem{Doshi-Velez2009UAI_correlated_latent}
Finale Doshi-Velez and Zoubin Ghahramani.
\newblock Correlated non-parametric latent feature models.
\newblock {\em UAI 2009}, 2009.

\bibitem{Hinton2006science_autoencoder}
G.~E. Hinton and R.~R. Salakhutdinov.
\newblock Reducing the dimension of data with neural networks.
\newblock {\em Science}, 313:504--507, 2006.

\bibitem{Hinton2002NeuralComputation_PoE_CD}
Geoffery~E. Hinton.
\newblock Training products of experts by minimizing contrastive divergence.
\newblock {\em Neural Computation}, 14:1771--1800, 2002.

\bibitem{Hinton1995Science_wake-sleep}
Geoffrey~E. Hinton, Peter Dayan, Brendan~J. Frey, and Neal~Radford M.
\newblock The "wake-sleep" algorithm for unsupervised neural networks.
\newblock {\em Science}, 268:1158--1161, 1995.

\bibitem{Hinton2006NeuralComputation_DBN}
Geoffrey~E. Hinton, S.~Osindero, and Yee-Whye Teh.
\newblock A fast learning algorithm for deep belief nets.
\newblock {\em Neural Computation}, 18:1527--1554, 2006.

\bibitem{Jaakkola1999JAIR_variational_QMR}
Tommi~S. Jaakkola and Michael~I. Jordan.
\newblock Variational probabilistic inference and the qmr-dt network.
\newblock {\em Journal of Artificial Intelligence Research}, 10:291--322, 1999.

\bibitem{LeCun_MNIST}
Yann LeCun, Corinna Cortes, and Christopher J.~C. Burges.
\newblock The mnist database.
\newblock {\em http://yann.lecun.com/exdb/mnist/}.

\bibitem{Obradovic1998NeuralComputation_infomax_ica}
D.~Obradovic and G.~Deco.
\newblock Information maximization and independent component analysis: Is there
  a difference?
\newblock {\em Neural Computation}, 10:2085--2101, 1998.

\bibitem{Ohata2003ICA_ICA_Helmholtz}
Masashi Ohata, Toshiharu Mukai, and Kiyotoshi Matsuoka.
\newblock Independent component analysis on the basis of helmholtz machine.
\newblock In {\em 4th International Symposium on Independent Component Analysis
  and Blind Signal Separation (ICA2003)}, 2003.

\bibitem{Pinchaud2011NIPS_mutual_information}
Nicolas Pinchaud.
\newblock Information theoretic learning of robust deep representations.
\newblock {\em NIPS 2011 Workshop on Deep Learning and Unsupervised Feature
  Learning}, 2011.

\bibitem{Quadrianto2013UAI_IBP}
Novi Quadrianto, Viktoriaa Sharmanska, David. KnKnowles, and Zoubin Ghahramani.
\newblock The supervised ibp: Neighbourhood preserving infinite latent feature
  models.
\newblock {\em Uncertainty in Artificial Intelligence UAI 2013}, 2013.

\bibitem{Salakhutdinov2007AISTATS_class_neighbourhood}
Ruslan Salakhutdinov and Geoffrey Hinton.
\newblock Learning a nonlinear embedding by preserving class neighbourhood
  structure.
\newblock {\em AI and Statistics (AISTATS) 2007}, 2007.

\bibitem{Salakhutdinov2007SIGIR_semantic_hashing}
Ruslan Salakhutdinov and Geoffrey Hinton.
\newblock Semantic hashing.
\newblock {\em International Journal of Approximate Reasoning}, 50(7):969--978,
  7 2007.

\bibitem{Shwe1990UAI_QMR}
Michael Shwe and Gregory~F. Cooper.
\newblock An empirical analysis of likelihood-weighting simulation on a large,
  multiply-connected belief network.
\newblock {\em Proceedings of the Sixth Conference on Uncertainty in Artificial
  Intelligence (UAI 1990)}, 1990.

\bibitem{Smolensky1986Book_harmony_theory}
Paul Smolensky.
\newblock {\em Parallel Distributed ProProcess: Explorations in the
  Microstructure of Cognition Volume 1}, volume~1.
\newblock MIT Press, 1986.

\bibitem{Theis2011JMLR_likelihood_db_not_enough}
Lucas Theis, Sebastian Gerwinn, Fabian Sinz, and Matthias Bethge.
\newblock In all likelihood, deep belief is not enough.
\newblock {\em Journal of Machine Learning Research}, 12:3071--3096, 2011.

\bibitem{Xu1998Neurocomputing_Ying-Yang}
Lei Xu.
\newblock Bayesian kullback ying-yang dependence reduction theory.
\newblock {\em Neurocomputing}, 22(1-3):81--111, 11 1998.

\end{thebibliography}






\end{document}